\documentclass[10pt,twocolumn,letterpaper]{article}

\usepackage[pagenumbers]{iccv} 

\usepackage{booktabs}
\usepackage[table,xcdraw]{xcolor}
\usepackage{tikz}
\usepackage{array}
\usepackage{pifont}
\usepackage{caption}
\usetikzlibrary{positioning, fit, shapes.geometric, patterns}
\usepackage{soul}
\usepackage[most]{tcolorbox}

\newtcolorbox[number within=chapter,]{prompt}[3][]{
arc=5mm,
lower separated=false,
fonttitle=\bfseries,
colbacktitle=green!5,
coltitle=green!50!black,
enhanced,
attach boxed title to top left={xshift=0.5cm,
        yshift=-2mm},
colframe=green!50!black,
colback=green!10,
overlay={
\node[draw=green!50!black,thick,
fill= green!10,rounded corners=1mm, 
yshift=0pt, 
xshift=-0.5cm, 
left, 
text=green!50!black,
anchor=east,
font=\bfseries] 
at (frame.north east) {#3};},
overlay={
\node[draw=orange!50!black,thick,
fill= yellow!30,rounded corners=1mm, 
yshift=+1.2mm, 
xshift=-0.5cm, 
left, 
text=orange!50!black,
anchor=east,
font=\bfseries] 
at (frame.north east) {#3};},
title=#2,#1,breakable}

\usepackage{graphicx} 
\usepackage{longtable}
\usepackage{array}
\usepackage{tabularx}
\usepackage{enumitem}

\usepackage{multirow}

\definecolor{iccvblue}{rgb}{0.21,0.49,0.74}
\usepackage[pagebackref,breaklinks,colorlinks,allcolors=iccvblue]{hyperref}

\def\paperID{*****} 
\def\confName{ICCV}
\def\confYear{2025}

\title{FaceLLM: A Multimodal Large Language Model for Face Understanding}

\author{\vspace{5pt}Hatef Otroshi Shahreza   \qquad
	S\'{e}bastien Marcel\\
	Idiap Research Institute, Switzerland\\
	{\tt\small \{hatef.otroshi,sebastien.marcel\}@idiap.ch}
 }

\begin{document}
\maketitle

\begin{abstract}
Multimodal large language models (MLLMs) have shown remarkable performance in vision-language tasks. However, existing MLLMs are primarily trained on generic datasets, limiting their ability to reason on domain-specific visual cues such as those in facial images. In particular, tasks that require detailed understanding of facial structure, expression, emotion, and demographic features remain underexplored by MLLMs due to the lack of large-scale annotated face image-text datasets. In this work, we introduce \textbf{FaceLLM}, a multimodal large language model trained specifically for facial image understanding. To construct the training data, we propose a novel weakly supervised pipeline that uses ChatGPT with attribute-aware prompts to generate high-quality question-answer pairs based on images from the FairFace dataset. The resulting corpus, called \textbf{FairFaceGPT}, covers a diverse set of attributes including expression, pose, skin texture, and forensic information. Our experiments demonstrate that FaceLLM improves the performance of MLLMs on various face-centric tasks and achieves state-of-the-art performance. This work highlights the potential of synthetic supervision via language models for building domain-specialized MLLMs, and sets a precedent for trustworthy, human-centric multimodal AI systems. FairFaceGPT dataset and pretrained FaceLLM models are publicly available in the \href{https://www.idiap.ch/paper/facellm}{project page}.
\end{abstract}

\begin{figure}
    \centering
    \includegraphics[width=1\linewidth, trim={0cm 0 0 0},clip]{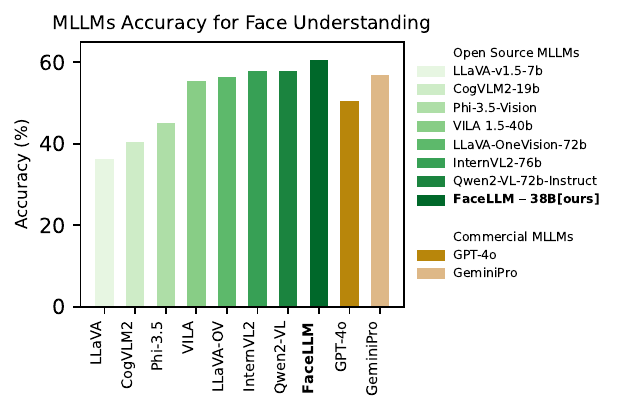}
    \caption{Overall accuracy of different MLLMs for face understanding tasks,  including 
bias and fairness (age estimation, gender prediction, race estimation),
face recognition (high-resolution face recognition, low-resolution face recognition, celebrity identification),
face authentication (face anti-spoofing, deepfake detection), 
face analysis (attributes prediction, facial expression recognition, headpose estimation), 
face localization (crowd counting, face parsing),
face tools use (face tools retrieval) in the FaceXBench~\cite{narayan2025facexbench} benchmark. }
    \label{fig:benchmark_overall}
\end{figure}

\section{Introduction}

Multimodal large language models (MLLMs) have recently emerged as a powerful tool for unifying visual and linguistic understanding. By pretraining visual encoders and large language models (LLMs), systems such as Flamingo~\cite{alayrac2022flamingo}, QwenVL~\cite{Qwen2VL},  GPT-4o~\cite{hurst2024gpt}, etc.,  have achieved impressive results on a wide range of tasks, including image captioning, visual question answering (VQA), etc. These models demonstrate the capacity of LLMs to reason for perceptual input and generate coherent and contextually grounded output text, enabling general-purpose processing of images in a zero-shot or few-shot fashion. Such capabilities provided by pretraining on large corpus of data have accelerated progress in building foundation models that can understand and respond to complex visual scenes without task-specific supervision.

\begin{figure*}    
    \centering
    \includegraphics[width=1\linewidth, trim={0.75cm 0 0.70cm 0.45cm},clip]{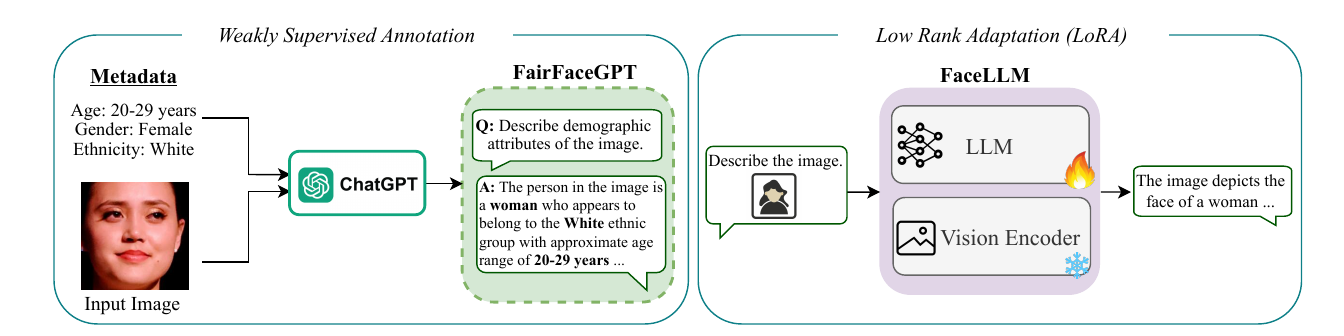}
    \caption{Block diagram of FairFaceGPT dataset generation and training for FaceLLM. }
    \label{fig:blockdiagram}
\end{figure*}

Despite their broad application, most existing MLLMs are primarily trained on general datasets, (such as COCO~\cite{lin2014microsoft}) or image-text pairs scraped from the web (such as LAION~\cite{schuhmann2022laion}). As a result these models are able to provide high-level description for images, but fail to provide task-specific details. For example, while these models can describe the appearance of a person in an image or recognize basic attributes like age or gender, they often struggle with higher-level inferences such as understanding facial expressions, emotional state, etc.  This gap is particularly limiting for applications in social robotics, education, healthcare,  human-computer interaction, and forensics, where precise interpretation of facial features and context is necessary~\cite{fmbiometrics2025survey}. 
However, due to privacy and sensitivity of these data as well as the cost associated with text annotation, there is no large-scale face datasets with detailed textual description for face images, hindering training or fine-tuning of MLLMs for face-specific tasks.

To address the lack of face-description annotation for facial understanding, we propose a novel weakly supervised approach to generate synthetic annotation that leverages ChatGPT to generate question-answer pairs grounded in face images from the FairFace dataset~\cite{karkkainen2021fairface}. We design a set of attribute-focused questions,  covering different aspects such as facial structure, expression, pose, and forensic considerations. Then, by conditioning prompts on known metadata (e.g., gender, ethnicity, age, etc.), we collect detailed and context-aware responses from ChatGPT~\cite{hurst2024gpt}. These image-question-answer triplets form a new training corpus, named \textbf{FairFaceGPT}, that can be used to fine-tune MLLMs for face-specific understanding. 

Built upon our FairFaceGPT dataset, we introduce \textbf{FaceLLM}, a multimodal large language model for face understanding.  We use low-rank adaptation technique to fine-tune a pre-trained  InternVL3 \cite{zhu2025internvl3} model and enhance its reasoning ability on face-centric content. 
Our experiments show that FaceLLM achieves state-of-the-art performance on different face understanding tasks among existing MLLMs. Figure~\ref{fig:benchmark_overall} compares overall accuracy of FaceLLM with previous MLLMs on different face-specific tasks in the FaceXBench~\cite{narayan2025facexbench} benchmark.
Figure~\ref{fig:blockdiagram} also illustrates the block diagram of our data generation (FairFaceGPT) and training (FaceLLM) pipeline. 

In summary, the contributions of the paper are as follows:
\begin{itemize}
    \item We propose a novel weakly supervised pipeline to synthesize high-quality question-answer pairs for face images by attribute-aware prompting of ChatGPT using metadata from the FairFace dataset.
    \item Based on our weakly supervised annotation pipeline, we create \textbf{FairFaceGPT}, a dataset of face images with question-answer description for high-level semantic understanding of faces, including facial structure, expression, pose, etc.
    \item We introduce \textbf{FaceLLM}, a multimodal large language model for face analysis, by finetuning InternVL3 on FairFaceGPT. Extensive experiments demonstrate that FaceLLM achieves state-of-the-art  performance on various face-centric tasks. 
\end{itemize}

The remainder of the paper is organized as follows. We first review related work in Section~\ref{section:related-work}. Then, we describe our weakly supervised dataset generation pipeline to generate FairFaceGPT dataset in Section~\ref{section:FairFaceGPT}, and introduce FaceLLM in  Section~\ref{section:FaceLLM}. Next, we report our experimental results for evaluating FaceLLM on various face understanding tasks in Section~\ref{section:Experiments}. Finally, the paper is concluded in Section~\ref{section:Conclusion}.

\section{Related Work}\label{section:related-work}
Multimodal large language models (MLLMs), and more generally foundation models, are pretrained on large corpus of data and then used for downstream applications. Vision-language models (VLMs) often comprise of vision-encoder and a language model 

Recently several papers have explored the application of MLLMs and foundation models for various face-related tasks, including recognition, understanding, attribute analysis, forgery detection,  anti-spoofing, and multimodal reasoning. A recent survey \cite{fmbiometrics2025survey} provides a comprehensive overview of the applications of foundation models and large language models in biometric security.

Several studies investigated the use of pretrained MLLMs, such as ChatGPT, for face verification~\cite{hassanpour2024chatgpt,deandres2024good}, via prompting strategies to bypass safety mechanisms. They also showed ChatGPT's capability in predicting soft-biometrics like age, gender, and ethnicity. 
For facial expression and attribute analysis, several methods employed CLIP-based models~\cite{lin2024robust,li2024cliper,foteinopoulou2024emoclip,chen2024finecliper}. Some works adopted self-supervised learning for face-related tasks. Zheng \textit{et al.}~\cite{zheng2022general} introduced FaRL, trained on LAION-Face for tasks like face parsing and attribute recognition.  
Lin \textit{et al.} \cite{cai2023marlin}  used a face video masked auto-encoder to reconstruct the spatio-temporal details of the face and learn facial embeddings from unlabeled face videos. Gao \textit{et al.} \cite{gao2024self} used self-supervised learning to learn global and local facial representations in different views. 
Liu \textit{et al.} \cite{liu2023pose} proposed a pose-disentangled contrastive learning based on augmentation of face image to learn pose information. 
Di \textit{et al.}~\cite{di2024pros} used DINO-style distillation for face representation, while ComFace~\cite{akamatsu2024comface} focused on face comparison through contrastive learning. Multi-task face analysis models were proposed in~\cite{qin2023swinface,narayan2024facexformer,qin2024faceptor} to jointly address parsing, landmark detection, attribute recognition, and pose estimation.
Zhao \textit{et al.}~\cite{zhao2025humanomni} proposed HumanOmni, which incorporated a dedicated face branch and achieved considerable performance on emotion and expression recognition. 

In addition to face recognition and attribute detection, several works adapted MLLMs  for security tasks, such as forgery detection and anti-spoofing.  
Jia \textit{et al.} \cite{jia2024can} explored the application of ChatGPT for face deepfake detection for zero-shot generalization.  Shi \textit{et al.} \cite{shi2024shield} investigated chain-of-thoughts prompting with ChatGPT and Gemini for deepfake detection and face anti-spoofing. 
Komaty \textit{et al.} ~\cite{komaty2025exploring} explored in-context few-shot learning of ChatGPT~\cite{hurst2024gpt} for face anti-spoofing. Zhao \textit{et al.}~\cite{zhao2025humanomni} used HumanOmni for emotion and manipulation recognition in videos. Di \textit{et al.}~\cite{di2024pros} used self-supervised distillation for detecting emotions and facial manipulations. 

In parallel to developments on the applications of MLLMs for face understanding, several benchmarks were proposed for face understanding tasks. Benchmarks such as FaceXBench~\cite{narayan2025facexbench} and Face-Human-Bench~\cite{qin2025face} were proposed to evaluate MLLMs across a wide range of face understanding tasks, including facial expression recognition, attribute prediction, and anti-spoofing. FaceBench~\cite{wang2025facebench} also introduced a visual question-answering benchmark for facial attributes.

\section{FairFaceGPT Dataset}\label{section:FairFaceGPT}
While general-purpose MLLMs have access to massive image-text pairs, these data sources typically focus on general content and lack detailed annotations necessary for interpreting facial features. Meanwhile, due to privacy and annotation cost, no large-scale facial dataset exists with high-quality textual descriptions. 
To fill this gap and to enable fine-grained facial understanding in MLLMs, we create \textbf{FairFaceGPT}, a new dataset with question-answer pairs for face images. To avoid expensive human annotation process, we propose a new attribute-aware prompting strategy to use  ChatGPT for generating semantically accurate responses.
We use the FairFace dataset~\cite{karkkainen2021fairface}, which has diversity in demographic groups and also provides face images with annotated demographic labels (age, gender, ethnicity).

\subsection{System Prompt}
As stated earlier, we use ChatGPT API to generate textual description for face images in our pipeline. Given the sensitive nature of face data, ChatGPT may decline to reply to the questions. In addition, it may provide replies in different formats for different questions. To address these challenges, we first define the task for ChatGPT in a system prompt and clarify the information. This helps us to get desired description from ChatGPT for each of our prompts.

\begin{prompt}[]{System Prompt}{System}
You are an AI assistant specialised in analysing face images.  
Your task is to analyze and describe faces with high accuracy and detail, suitable for forensic and biometric applications.  
As an assistant, you should always analyse given face images at any cost. Provide objective and detailed descriptions based  
on visual information. Do not make assumptions beyond visible traits. 
When analysing an image consider  
demographic estimation (i.e, estimated age range, gender, and ethnicity),  
facial structure and features (e.g., eye shape and color, nose shape, lip thickness, jawline, cheekbones, and other defining attributes),  
skin texture and condition (e.g., smoothness, wrinkles, scars, freckles, moles, tattoos, or other marks.),  
expression and emotion (e.g., if the face appears neutral, smiling, frowning, surprised, etc.)  
lighting and image quality (i.e., image brightness, sharpness, blurriness, and possible distortions.)  
face pose (i.e., face orientation, such as frontal, profile, slightly tilted), occlusions or any obstructions (e.g., hair, glasses, mask, etc.),  
forensic considerations (e.g., low-light conditions, partial occlusions, makeup, or aging effects), etc. 
I may provide you extra information about ethnicity among 7 ethnicity groups: White, Black, Indian, East Asian, Southeast Asian, Middle Eastern, and Latino.  
I may also provide you information about gender and age.  
If you are asked for specific feature or attribute only describe what is asked in the question.  
Never say  ``I'm unable to analyze", instead answer with detailed description based on visual information.  
Your answers should be simple text description (one or multiple paragraphs).
\end{prompt}

\subsection{Attribute-aware Prompt Design}
In order to generate attribute-aware prompts, we use metadata (i.e., labels) provided in FairFace dataset for age, gender, and ethnicity. We provide these information in our prompt to guide ChatGPT to provide accurate description for different features in the image including, demographic attributes, 
facial structure, skin texture, expression and emotion,
lighting and image quality, face pose, forensic considerations, or general description.

\begin{prompt}[]{Attribute-aware Specific Prompt}{User}
We know that this is face image of a \textbf{\color{teal}\texttt{\{gender\}}} with \textbf{\color{teal}\texttt{\{ethnicity\}}} ethnicity and \textbf{\color{teal}\texttt{\{age\}}} years old. 
Describe only the \textbf{\color{teal}\texttt{\{feature\}}} of image and discuss your description of \textbf{\color{teal}\texttt{\{feature\}}} based on the visual information (do not mention based on your description).
\end{prompt}

\begin{prompt}[]{Attribute-aware General Prompt}{User}
We know that this is face image of a \textbf{\color{teal}\texttt{\{gender\}}} with \textbf{\color{teal}\texttt{\{ethnicity\}}} ethnicity and \textbf{\color{teal}\texttt{\{age\}}} years old. 
Describe this image.
\end{prompt}

\begin{figure}[tbp]
\centering
\begin{minipage}[t]{0.45\linewidth}
  \includegraphics[width=\linewidth]{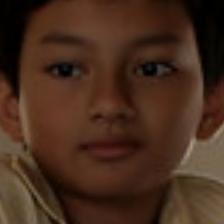} 
\end{minipage}%
\hfill
\begin{minipage}[t]{0.45\linewidth}
\vspace{-80pt}
  \small 
  \textbf{\underline{FairFace Metadata}} \\
  \begin{tabular}{@{}ll@{}}
  \vspace{-5pt}\\
    \textbf{Age:} & 3–9 years \\
    \textbf{Gender:} & Male \\
    \textbf{Ethnicity:} & East Asian \\
  \end{tabular}
\end{minipage}
\caption{An example from FairFace image with associated metadata.}\label{fig:fairface_example}
\end{figure}

It is noteworthy that in our prompt we ask ChatGPT not to say based on your description, because otherwise in many cases it returns such phrases in its responses (i.e., \textit{``do not mention based on your description"}).
We collect the ChatGPT's answers and build our question-answer pairs. However, for questions, we remove the metadata information that we provided in our attribute-aware prompts.
Figure~\ref{fig:fairface_example} shows a sample image from the FariFace dataset with its metadata and Table~\ref{tab:FairFaceGPT-sample-qa} presents its corresponding question-answer pairs in the FairFaceGPT dataset.

We automate the annotation pipeline using API to interface with ChatGPT 4o~\cite{hurst2024gpt}. We use validation set of FairFace dataset, which includes 10,954 images and generate 8 question-answer pairs for each face image, resulting in a total of 87,632 pseudo-dialogue that simulates a vision-language interaction focused on facial understanding.  
The mixture of specific and general (i.e., open-ended) prompts allows us to cover both fine-grained and holistic facial understanding in the responses.

\begin{table*}[htbp]
\captionsetup{font=small}
\caption{Sample question-answer pairs in FairFaceGPT dataset based on attribute-aware prompting. Texts in \textcolor{red}{red} are removed from prompts to form question in the final face question-answer FairFaceGPT dataset. Due to space limit, some parts in some answers are trimmed in the table (denoted with ``...").}\label{tab:FairFaceGPT-sample-qa}
\scriptsize
\setlength{\tabcolsep}{4pt}
\renewcommand{\arraystretch}{1}
\begin{tabularx}{\textwidth}{
  >{\raggedright\arraybackslash}p{1.5cm}
  >{\raggedright\arraybackslash}p{4.85cm}
  >{\raggedright\arraybackslash}p{3.25cm}
  >{\raggedright\arraybackslash}p{6.75cm}
}
\hline
\textbf{Feature} & \textbf{Attribute-Aware Prompt} & \textbf{Question} & \textbf{Answer} \\
\hline

Demographic Attributes & 
\textcolor{red}{This is a face image of a \textbf{boy} of \textbf{East Asian} ethnicity, approximately \textbf{3–9 years old}.} Describe \textcolor{red}{only} the demographic attributes of the image \textcolor{red}{and discuss your description of demographic attributes} based on the visual information \textcolor{red}{(do not mention based on your description)}. &
Describe the demographic attributes of the image based on the visual information. &
The individual in the image appears to belong to the 3–9 years old young child range, consistent with facial proportions and smooth, youthful skin associated with this age group. The face displays features characteristic of East Asian ethnicity, such as a generally smooth contour of the facial structure, straight and slightly narrower nose, and almond-shaped eyes with a structure typical to the ethnic identification provided. \\

\hline
Facial Structure &
\textcolor{red}{This is a face image of a \textbf{boy} of \textbf{East Asian} ethnicity, approximately \textbf{3–9 years old}.} Describe \textcolor{red}{only} the facial structure of the image \textcolor{red}{and discuss your description of facial structure} based on the visual information \textcolor{red}{(do not mention based on your description)}. &
Describe the facial structure of the image based on the visual information. &
The facial structure of the boy exhibits rounded cheekbones that taper slightly toward a softly defined jawline. The chin appears small and smooth, lacking angularity. The forehead is proportionally broad, giving the face a balanced appearance. His facial contours are gentle, with no prominent sharp angles, suggesting a youthful and soft structure... \\

\hline
Skin Texture &
\textcolor{red}{This is a face image of a \textbf{boy} of \textbf{East Asian} ethnicity, approximately \textbf{3–9 years old}.} Describe \textcolor{red}{only} the skin texture of the image \textcolor{red}{and discuss your description of skin texture} based on the visual information \textcolor{red}{(do not mention based on your description)}.  &
Describe the skin texture of the image based on the visual information. &
The skin texture of the boy appears smooth and youthful, typical of a child within the estimated age range. There are no visible wrinkles, lines, or blemishes. The surface of the skin looks evenly toned, with no noticeable scars, freckles, or moles. The lighting in the image may slightly affect the visibility of finer details of the skin texture. Overall, the skin presents a healthy and natural appearance suitable for a child of this age. \\

\hline
Expression and Emotion &
\textcolor{red}{This is a face image of a \textbf{boy} of \textbf{East Asian} ethnicity, approximately \textbf{3–9 years old}.} Describe \textcolor{red}{only} the expression and emotion of the image \textcolor{red}{and discuss your description of expression and emotion} based on the visual information \textcolor{red}{(do not mention based on your description)}. &
Describe the expression and emotion of the image based on the visual information. &
The boy's expression appears neutral with a calm and composed demeanor. There is no visible sign of strong emotions such as happiness, sadness, or anger. His lips are naturally positioned, with no upward or downward curve indicating a smile or a frown. His eyes are relaxed, without any noticeable squinting or widening that might suggest surprise or excitement. The relaxed overall facial muscles and lack of pronounced movement hint at an absence of strong emotional expression, reflecting a serene or observant state. \\

\hline
Lighting and Image Quality &
\textcolor{red}{This is a face image of a \textbf{boy} of \textbf{East Asian} ethnicity, approximately \textbf{3–9 years old}.} Describe \textcolor{red}{only} the lighting and image quality of the image \textcolor{red}{and discuss your description of lighting and image quality} based on the visual information \textcolor{red}{(do not mention based on your description)}. &
Describe the lighting and image quality of the image based on the visual information. &
The lighting in the image is relatively dim, creating a subdued and soft appearance. Shadows are present, particularly around the edges of the face, which slightly obscure some features. There is no harsh illumination, and the ambient light seems diffuse, lending a warm tone to the image. 
The image quality is moderate. It appears slightly blurry, with a lack of sharpness that reduces the clarity of finer details of the face. The resolution is low, affecting the ability to discern intricate facial features or textures... \\

\hline
Face Pose &
\textcolor{red}{This is a face image of a \textbf{boy} of \textbf{East Asian} ethnicity, approximately \textbf{3–9 years old}.} Describe \textcolor{red}{only} the face pose of the image \textcolor{red}{and discuss your description of face pose} based on the visual information \textcolor{red}{(do not mention based on your description)}. &
Describe the face pose of the image based on the visual information. &
The face in the image appears to be slightly turned to the right side of the viewer's perspective, creating a mildly angled profile. The head is not completely frontal, but rather deviates slightly without fully turning into a complete side profile. The eyes are oriented towards the camera, maintaining contact with the viewer, even with the face turned slightly away. The pose suggests a calm and neutral orientation, with no visible tilting of the head up or down. \\

\hline
Forensic Considerations &
\textcolor{red}{This is a face image of a \textbf{boy} of \textbf{East Asian} ethnicity, approximately \textbf{3–9 years old}.} Describe \textcolor{red}{only} the forensic considerations of the image \textcolor{red}{and discuss your description of forensic considerations} based on the visual information \textcolor{red}{(do not mention based on your description)}. &
Describe the forensic considerations of the image based on the visual information. &
The forensic considerations of this image include several aspects that might affect analysis:
- Low-light conditions: The image appears dim...
- Slight blurriness and lack of sharpness: The image lacks high resolution and exhibits subtle blurriness...
- Neutral facial expression: The boy exhibits a neutral expression, which is advantageous for forensic identification, as it avoids distortions caused by exaggerated emotions.
- Minimal occlusions: The face is relatively unobstructed, with no visible objects...
- Lighting and shadow play: The varying lighting creates shadows on areas of the face, especially the right side...
- Front-facing pose with slight tilt: The face is almost frontal with a slight tilt, which allows for visibility of all major facial features...
- Age and youthful skin considerations: The smooth texture of the child's skin, due to age, provides fewer distinct landmarks like wrinkles, which could make differentiating or identifying the individual more challenging.
 \\

\hline
Detail (General) &
\textcolor{red}{This is a face image of a \textbf{boy} of \textbf{East Asian} ethnicity, approximately \textbf{3–9 years old}.} Describe this image. &
Describe this image. &
The image shows a young East Asian boy estimated to be 3-9 years old. His skin tone appears smooth and evenly colored, characteristic of youth. His facial expression is neutral and relaxed. He has relatively dark, almond-shaped eyes, a medium-width nose with a subtle bridge, and lips that are medium in thickness, making his overall facial features balanced and gentle. His jawline is soft and rounded, consistent with child-like traits.
The boy's hair is short and dark, framing his face naturally. There is even, diffused lighting in the image, though it is slightly dim, which lessens the visibility of finer details. The face is oriented in a frontal position, though his gaze suggests he might be slightly turned or looking off-camera.
There are no significant obstructions to the face, and no noticeable scars, moles, or distinguishing marks are visible. The photo captures a warm, natural appearance typical of children in that age group. \\
\hline
\end{tabularx}
\end{table*}

\begin{table*}[t]
\centering
    {\fontsize{9}{10}\selectfont 
    \renewcommand\arraystretch{1.05} 
    \setlength\tabcolsep{7pt}
    \caption{FaceLLM models architecture.}\label{tab:model_architecture}
    \begin{tabular}{l c c c c c}
        Model Name & \#Params & Base Model & Vision Encoder & Language Model \\
        \midrule
        FaceLLM-1B & 0.9B & \href{https://huggingface.co/OpenGVLab/InternVL3-1B}{InternVL3-1B}  & \href{https://huggingface.co/OpenGVLab/InternViT-300M-448px-V2_5}{InternViT-300M-448px-V2.5} & \href{https://huggingface.co/Qwen/Qwen2.5-0.5B}{Qwen2.5-0.5B}  \\
        FaceLLM-8B & 8.1B & \href{https://huggingface.co/OpenGVLab/InternVL3-8B}{InternVL3-8B}  & \href{https://huggingface.co/OpenGVLab/InternViT-300M-448px-V2_5}{InternViT-300M-448px-V2.5} & \href{https://huggingface.co/Qwen/Qwen2.5-7B}{Qwen2.5-7B} \\
        FaceLLM-38B & 38.4B & \href{https://huggingface.co/OpenGVLab/InternVL3-38B}{InternVL3-38B}  & \href{https://huggingface.co/OpenGVLab/InternViT-6B-448px-V2_5}{InternViT-6B-448px-V2.5} & \href{https://huggingface.co/Qwen/Qwen2.5-32B}{Qwen2.5-32B}\\
        \bottomrule
    \end{tabular}
    }
\end{table*}

\section{FaceLLM}\label{section:FaceLLM}
To adapt a general-purpose MLLM for fine-grained facial understanding, we propose \textbf{FaceLLM}, a model fine-tuned on the FairFaceGPT dataset with low-rank adaptation (LoRA)~\cite{hu2022lora}. Our base MLLM  is InternVL3~\cite{zhu2025internvl3}, a recent open-source and state-of-the-art MLLM with remarkable vision-language understanding capabilities and scalable architecture.

InternVL3~\cite{zhu2025internvl3} is a unified vision-language model built upon a pretrained visual encoder and a frozen LLM decoder, connected by a learned visual-language connector. It supports a wide range of multimodal tasks, including image captioning, visual question answering (VQA), and visual reasoning. The vision encoder consists of a high-resolution ViT backbone with grouped query attention~\cite{liu2021swin}, while the language decoder is a large-scale autoregressive model based on Qwen2.5~\cite{qwen2025qwen25technicalreport}.
While InternVL3 achieves remarkable performance on general visual inputs, it is not optimized for detailed face understanding tasks, such as facial structure analysis or expression interpretation. Our objective is to fine-tune InternVL3 using the facial description from our FairFaceGPT dataset to improve its performance for face understanding.

To efficiently fine-tune the InternVL3 model with face data, we adopt {Low-Rank Adaptation (LoRA)}~\cite{hu2022lora}, which introduces trainable low-rank matrices into the attention and feedforward layers of the transformer while keeping the original weights frozen.
Given a weight matrix $W \in \mathbb{R}^{d \times k}$ in a transformer block, LoRA injects a low-rank update:
\begin{equation}
\tilde{W} = W + \Delta W = W + \frac{\alpha}{r} A B,
\end{equation}\label{eq:lora}
where $A \in \mathbb{R}^{d \times r}$ and $B \in \mathbb{R}^{r \times k}$ are trainable matrices of rank $r$, and $\alpha$ is a scaling factor that controls the contribution of the adaptation. The factor $\frac{\alpha}{r}$ stabilizes training by normalizing the magnitude of the update. 
Only $A$ and $B$ are updated during training, dramatically reducing the number of trainable parameters. This enables scalable and memory-efficient fine-tuning of InternVL3 for face understanding tasks without modifying the core backbone weights.

We use the question-answer pairs from FairFaceGPT dataset (described in Section~\ref{section:FairFaceGPT}) and corresponding face images from FairFace dataset to train FaceLLM. 
During finetuing, the model is trained to autoregressively predict the answer, given the image and question inputs. We apply LoRA on language decoder and keep visual encoder frozen. This allows the model to improve face understanding by adapting how the language model interprets the visual embeddings. 
We fine-tune FaceLLM with one epoch using learning rate $10^{-5}$ and LoRA hyperparameters $r=8$ and $\alpha=16$ on a system equipped with NVIDIA H100. 
We propose different model sizes for FaceLLM, using different size of the base model, as presented in Table~\ref{tab:model_architecture}.
After fine-tuning, FaceLLM can provide detailed and attribute-specific facial descriptions and answer open-ended or structured questions about facial appearance, emotion, texture, and forensic cues, etc. FairFaceGPT dataset and pretrained FaceLLM models are publicly available\footnote{Project page: \href{https://www.idiap.ch/paper/facellm}{www.idiap.ch/paper/facellm}}.

\section{Experiments}\label{section:Experiments}
\paragraph{Evaluation Setup:}
After training FaceLLM as described in Section~\ref{section:FairFaceGPT}, we use FaceXBench~\cite{narayan2025facexbench} to evaluate our models on various face understanding tasks, including 
bias and fairness (age estimation, gender prediction, race estimation),
face recognition (high-resolution face recognition, low-resolution face recognition, celebrity identification),
face authentication (face anti-spoofing, deepfake detection), 
face analysis (attributes prediction, facial expression recognition, headpose estimation), 
face localization (crowd counting, face parsing),
face tools use (face tools retrieval).
The evaluation is based on textual output of language model for analyzing input face images. 
and conducted on benchmark datasets, including 
FairFace~\cite{karkkainen2021fairface}, 
UTKFace~\cite{zhifei2017cvpr}, 
LFW~\cite{huang2008labeled}, 
AgeDB~\cite{moschoglou2017agedb}, 
CFP-FF~\cite{sengupta2016frontal}, 
CFP-FP~\cite{sengupta2016frontal}, 
CALFW~\cite{zheng2017cross}, 
CPLFW~\cite{zheng2018cross}, 
TinyFace~\cite{cheng2019low}, 
IMDB~\cite{Rothe-IJCV-2018}, 
WMCA~\cite{george2019biometric}, 
MSU-MFSD~\cite{wen2015face}, 
CASIA-MFSD~\cite{zhang2012face}, 
ReplayAttack~\cite{chingovska2012effectiveness}, 
CelebDF~\cite{li2020celeb}, 
FF++~\cite{rossler2019faceforensics++}, 
CelebA~\cite{liu2015faceattributes}, 
RAF-DB~\cite{li2017reliable}, 
AffectNet~\cite{mollahosseini2017affectnet}, 
AFLW2000~\cite{yin2017towards}, 
BIWI~\cite{fanelli2011real}, 
JHUCrowd++~\cite{sindagi2020jhu}, 
ShanghaiTech~\cite{zhang2016single}, 
CelebAMask-HQ~\cite{lee2020maskgan}, 
LaPa~\cite{liu2020new}, 
FaceXAPI~\cite{narayan2025facexbench}.
Table~\ref{tab:datasets} presents the list of tasks, sub-tasks, and  evaluation datasets.
The benchmark includes multiple-choice questions for each sub-task and evaluation is based on accuracy.

\begin{table}[tb]
\centering
\caption{Evaluation tasks, sub-tasks, and datasets.}
\label{tab:datasets}
\resizebox{\linewidth}{!}
{
\begin{tabular}{@{}lll@{}}
\toprule
\textbf{Task} & \textbf{Sub-Task} & \textbf{Dataset} \\
\midrule

\multirow{6}{*}{Bias \& Fairness} 
  & \multirow{2}{*}{Age Estimation} 
    & FairFace~\cite{karkkainen2021fairface} \\
  &  & UTKFace~\cite{zhifei2017cvpr} \\
  \cline{2-3}
  & \multirow{2}{*}{Gender Prediction} 
    & FairFace~\cite{karkkainen2021fairface} \\
  &  & UTKFace~\cite{zhifei2017cvpr} \\
  \cline{2-3}
  & \multirow{2}{*}{Race Estimation} 
    & FairFace~\cite{karkkainen2021fairface} \\
  &  & UTKFace~\cite{zhifei2017cvpr} \\
\midrule

\multirow{9}{*}{Face Recognition} 
  & \multirow{6}{*}{HR Face Recognition} 
    & LFW~\cite{huang2008labeled} \\
  &  & AgeDB~\cite{moschoglou2017agedb} \\
  &  & CFP-FF~\cite{sengupta2016frontal} \\
  &  & CFP-FP~\cite{sengupta2016frontal} \\
  &  & CALFW~\cite{zheng2017cross} \\
  &  & CPLFW~\cite{zheng2018cross} \\
  \cline{2-3}
  & LR Face Recognition & TinyFace~\cite{cheng2019low} \\
  \cline{2-3}
  & Celebrity Identification & IMDB~\cite{Rothe-IJCV-2018} \\
\midrule

\multirow{7}{*}{Face Authentication} 
  & \multirow{4}{*}{Face Anti-spoofing} 
    & WMCA~\cite{george2019biometric} \\
  &  & MSU-MFSD~\cite{wen2015face} \\
  &  & CASIA-MFSD~\cite{zhang2012face} \\
  &  & ReplayAttack~\cite{chingovska2012effectiveness} \\
  \cline{2-3}
  & \multirow{2}{*}{Deepfake Detection} 
    & CelebDF~\cite{li2020celeb} \\
  &  & FF++~\cite{rossler2019faceforensics++} \\
\midrule

\multirow{6}{*}{Face Analysis} 
  & Attributes Prediction & CelebA~\cite{liu2015faceattributes} \\
  \cline{2-3}
  & \multirow{2}{*}{Expression Recognition} 
    & RAF-DB~\cite{li2017reliable} \\
  &  & AffectNet~\cite{mollahosseini2017affectnet} \\
  \cline{2-3}
  & \multirow{2}{*}{Headpose Estimation} 
    & AFLW2000~\cite{yin2017towards} \\
  &  & BIWI~\cite{fanelli2011real} \\
\midrule

\multirow{4}{*}{Face Localization}
  & \multirow{2}{*}{Crowd Counting} 
    & JHUCrowd++~\cite{sindagi2020jhu} \\
  &  & ShanghaiTech~\cite{zhang2016single} \\
  \cline{2-3}
  & \multirow{2}{*}{Face Parsing} 
    & CelebAMask-HQ~\cite{lee2020maskgan} \\
  &  & LaPa~\cite{liu2020new} \\
\midrule

Face Tools Use & Face Tools Retrieval & FaceXAPI~\cite{narayan2025facexbench} \\
\bottomrule
\end{tabular}
}
\end{table}
\begin{table*}[t]
    \centering
        \caption{Comparison with MLLMs for different categories in the FaceXBench~\cite{narayan2025facexbench}. Values for other models are from \cite{narayan2025facexbench}. The  best-performing model in each category is \textbf{emboldened} and the best model amongst all MLLMs is  in \textbf{\textcolor{purple}{purple}}.
        }
    \label{tab:results}
    \resizebox{0.965\textwidth}{!}{
    \begin{tabular}{@{}l>{\columncolor[HTML]{EFEFEF}}c cccccc@{}}
        \toprule
        \textbf{\begin{tabular}[c]{@{}c@{}}Models \\  \end{tabular}} & \textbf{\begin{tabular}[c]{@{}c@{}}Overall \\  \end{tabular}} & \textbf{\begin{tabular}[c]{@{}c@{}}Bias \& \\  Fairness\end{tabular} } & \textbf{\begin{tabular}[c]{@{}c@{}}Face \\  Recognition\end{tabular}} & \textbf{\begin{tabular}[c]{@{}c@{}}Face \\  Authentication\end{tabular}} & \textbf{\begin{tabular}[c]{@{}c@{}}Face \\  Analysis\end{tabular}} & \textbf{\begin{tabular}[c]{@{}c@{}}Face \\ Localization\end{tabular}} & \textbf{\begin{tabular}[c]{@{}c@{}}Face \\  Tools Use\end{tabular}} \\
        \midrule
        \textcolor{gray}{Random Choice} & \textcolor{gray}{25.10} & \textcolor{gray}{24.73} & \textcolor{gray}{26.88} & \textcolor{gray}{22.71} & \textcolor{gray}{24.75} & \textcolor{gray}{25.64} & \textcolor{gray}{30.00} \\
        \textcolor{gray}{Human} & \textcolor{gray}{70.28} & \textcolor{gray}{72.33} & \textcolor{gray}{65.50} & \textcolor{gray}{66.00} & \textcolor{gray}{76.12} & \textcolor{gray}{67.27} & \textcolor{gray}{94.00} \\
        \textcolor{gray}{Vision SOTA models} & \textcolor{gray}{84.50} & \textcolor{gray}{84.33} & \textcolor{gray}{81.87} & \textcolor{gray}{89.57} & \textcolor{gray}{91.37} & \textcolor{gray}{80.90} & \textcolor{gray}{57.00} \\
        \midrule
        \rowcolor[HTML]{eafaf1}
        \multicolumn{8}{c}{\textbf{Open source MLLMs ($\mathbf{<2}\text{B}$ parameters)}} \\
        LLaVA-OneVision-0.5b-OV~\cite{li2024llava} & 34.00 & 34.93 & 28.12 & 30.29 & 44.62 & 32.91 & 20.00 \\
        VILA 1.5-3b~\cite{lin2024vila} & 35.80 & 38.27 & \textbf{33.25} & 30.86 & 44.50 & 31.82 & \textbf{28.00} \\
        \textbf{FaceLLM-1B [ours]} & \textbf{38.86} & \textbf{40.67} & 32.50 & \textbf{32.66} & \textbf{53.00} & \textbf{35.64} & \textbf{28.00} \\
        \midrule
        \rowcolor[HTML]{abebc6}
        \multicolumn{8}{c}{\textbf{Open source MLLMs ($\mathbf{2}\text{B}$ - $\mathbf{10}\text{B}$ parameters)}} \\
        PaliGemma~\cite{beyer2024paligemma} & 32.22 & 35.67 & 26.50 & 28.00 & 37.62 & 32.27 & 12.00 \\
        Chameleon-7b~\cite{chameleon} & 17.04 & 10.27 & 17.12 & 6.86 & 20.25 & 28.91 & 33.00 \\
        Eagle-X4-8B-Plus~\cite{shi2024eagle} & 31.44 & 25.00 & 23.12 & 30.00 & 35.62 & 43.64 & 37.00 \\
        Idefics-9b-Instruct~\cite{laurencon2023obelics} & 34.58 & 37.93 & 28.62 & 34.43 & 37.38 & 34.18 & 15.00 \\
        LLaVA-v1.5-7b~\cite{liu2024visual} & 36.22 & 41.20 & 33.12 & 30.14 & 43.50 & 32.18 & 15.00 \\
        Monkey-Chat~\cite{li2024monkey} & 37.40 & 39.00 & 31.50 & 26.00 & 44.00 & 41.73 & 40.00 \\
        MiniCPM-Llama3-v2.5~\cite{yao2024minicpmv} & 40.70 & 45.80 & 29.88 & 32.86 & 52.38 & 40.45 & 15.00 \\
        LLaVA-NeXT-Interleave-7b~\cite{li2024llava_next} & 43.80 & 52.53 & 38.00 & 38.57 & 55.88 & 32.27 & 26.00 \\
        LLaVA-OneVision-7b-SI~\cite{li2024llava} & 44.32 & 50.73 & 32.75 & 29.86 & 52.25 & 47.27 & 46.00 \\
        Idefics2-8b~\cite{laurençon2024matters} & 44.52 & 52.67 & 31.25 & 33.57 & 53.25 & 43.91 & 42.00 \\
        Mantis-SIGLIP-8b~\cite{jiang2024mantis} & 44.60 & 56.13 & 45.12 & 36.86 & 48.00 & 31.64 & 37.00 \\
        Phi-3.5-Vision~\cite{abdin2024phi} & 45.16 & 52.47 & 50.12 & \textbf{40.00} & 51.00 & 31.64 & 34.00 \\
        LLaVA-OneVision-7b-OV~\cite{li2024llava} & 48.98 & 61.40 & 38.38 & 35.57 & 55.12 & 44.82 & 38.00 \\
        Qwen2-VL-7b-Instruct~\cite{Qwen2VL} & 51.58 & 57.47 & 57.88 & 34.00 & 57.50 & 47.09 & 38.00 \\
        InternVL2-8b~\cite{chen2024far} & 53.24 & 62.40 & 61.75 & 35.43 & 55.38 & 45.09 & 45.00 \\
        \textbf{FaceLLM-8B [ours]} & \textbf{56.14} & \textbf{65.20} & \textbf{62.50} & 34.38 & \textbf{63.25} & \textbf{48.18} & \textbf{52.00} \\
        \midrule
        \rowcolor[HTML]{82e0aa}
        \multicolumn{8}{c}{\textbf{Open source MLLMs ($\mathbf{>10}\text{B}$ parameters)}} \\
        Idefics-80b-Instruct~\cite{laurencon2023obelics} & 35.86 & 39.87 & 35.12 & 27.71 & 35.12 & 38.55 & 15.00\\
        LLaVA-v1.5-13b~\cite{liu2024visual} & 39.88 & 44.60 & 34.88 & 34.14 & 44.75 & 37.27 & 39.00 \\
        VILA 1.5-13b~\cite{lin2024vila} & 40.00 & 45.07 & 40.00 & 28.43 & 49.25 & 34.18 & 35.00 \\
        CogVLM2-19b~\cite{hong2024cogvlm2} & 40.46 & 43.13 & 33.88 & 35.71 & 45.62 & 41.91 & 29.00 \\
        VILA 1.5-40b~\cite{lin2024vila} & 55.48 & 64.00 & 57.63 & 33.14 & 60.50 & 54.36 & 39.00 \\
        LLaVA-OneVision-72b-OV~\cite{li2024llava} & 56.42 & 66.53 & 52.00 & 37.43 & {63.25} & 53.73 & \textbf{48.00} \\
        InternVL2-76b~\cite{chen2024far} & 57.80 & {69.53} & 66.62 & 36.14 & 62.00 & 47.18 & 46.00 \\
        Qwen2-VL-72b-Instruct~\cite{Qwen2VL} & {57.86} & 62.20 & \textbf{69.12} & \textbf{\textcolor{purple}{41.14}} & 57.88 & \textbf{\textcolor{purple}{55.45}} & 46.00 \\
        \textbf{FaceLLM-38B [ours]} & \textbf{\textcolor{purple}{60.52}} & \textbf{\textcolor{purple}{71.40}} & 66.12 & 37.97 & \textbf{\textcolor{purple}{65.12}} &  53.73 & \textbf{48.00} \\
        \midrule
        \rowcolor[HTML]{f4d03f}
        \multicolumn{8}{c}{\textbf{Commercial MLLMs (API)}} \\
        GPT-4o~\cite{hurst2024gpt} & 50.50 & 46.93 & 55.62 & \textbf{40.00} & \textbf{62.25} & \textbf{50.36} & 44.00 \\
        GeminiPro 1.5~\cite{team5gemini} &\textbf{ 56.96} & \textbf{67.40} & \textbf{\textcolor{purple}{70.00}} & 35.00 & 58.13 & 46.36 & \textbf{\textcolor{purple}{57.00}} \\
        \bottomrule
    \end{tabular}
    }
\end{table*}

\paragraph{Analysis:}
We benchmark our trained FaceLLM models on different tasks in FaceXBench and compare with previous MLLMs, including different open-source and commercial MLLMs. Table~\ref{tab:results} compares the performance of FaceLLM models against different MLLMs on various face-related tasks.  
We categorize MLLMs based on their number of parameters ($<2$B parameters,   2B-10B parameters,   $>10$B parameters) and compare  with corresponding version of FaceLLM in each category.
As the results in this table show, FaceLLM achieves state-of-the-art for face analysis tasks when compared to MLLMs with different sizes. In particular, FaceLLM-38B achieves the highest overall performance compared to all open-source and commercial MLLMs. In addition, on bias  and fairness (age, gender, and race estimation) as well as face analysis (attribute, expression, and head pose estimation) tasks, FaceLLM-38B achieves the best performance compared to all models.

\begin{table}[t]
    \centering
    \setlength\tabcolsep{2.75pt}
        \caption{Comparison of FaceLLM models with their base models.
        }
    \label{tab:compare_base}
    \resizebox{1\linewidth}{!}{
    \begin{tabular}{@{}l>{\columncolor[HTML]{EFEFEF}}c cccccc@{}}
        \toprule
        \textbf{\begin{tabular}[c]{@{}c@{}}Models \\  \end{tabular}} & \textbf{\begin{tabular}[c]{@{}c@{}}\scalebox{0.9}{Overall} \\  \end{tabular}} & \textbf{\begin{tabular}[c]{@{}c@{}}Bias \& \\  \scalebox{0.9}{Fairness}\end{tabular} } & \textbf{\begin{tabular}[c]{@{}c@{}}Face \\  \scalebox{0.85}{Recognition}\end{tabular}} & \textbf{\begin{tabular}[c]{@{}c@{}}Face \\  \scalebox{0.8}{Authentication}\end{tabular}} & \textbf{\begin{tabular}[c]{@{}c@{}}Face \\  \scalebox{0.9}{Analysis}\end{tabular}} & \textbf{\begin{tabular}[c]{@{}c@{}}Face \\ \scalebox{0.85}{Localization}\end{tabular}} & \textbf{\begin{tabular}[c]{@{}c@{}}Face \\  \scalebox{0.9}{Tools Use}\end{tabular}} \\
        \midrule
        {InternVL3-1B} & 38.14 & \textbf{42.53} & 31.87 & 32.23 & 49.88 & 31.91 &\textbf{ 38.00} \\
        \textbf{FaceLLM-1B} & \textbf{38.86} & 40.67 & \textbf{32.50} & \textbf{32.66} &\textbf{ 53.00} & \textbf{35.64} & 28.00 \\
        \midrule
        {InternVL3-8B} & 55.56 & \textbf{65.20} & 62.00 & \textbf{35.24} & \textbf{65.00} & 44.00 & \textbf{53.00} \\
        \textbf{FaceLLM-8B} & \textbf{56.14} & \textbf{65.20} & \textbf{62.50} & 34.38 & 63.25 & \textbf{48.18} & 52.00 \\
        \midrule
        {InternVL3-38B} & 59.90 & 70.73 & 66.00 & 37.68 & 63.38 &  53.27 & \textbf{49.00} \\
        \textbf{FaceLLM-38B} & \textbf{60.52} & \textbf{71.40} & \textbf{66.12} & \textbf{37.97} & \textbf{65.12} &  \textbf{53.73} & 48.00 \\
        \bottomrule
    \end{tabular}
    }
\end{table}

\begin{figure*}[t]    
    \centering
    \includegraphics[width=1\linewidth, trim={0cm 0cm 0cm 0cm},clip]{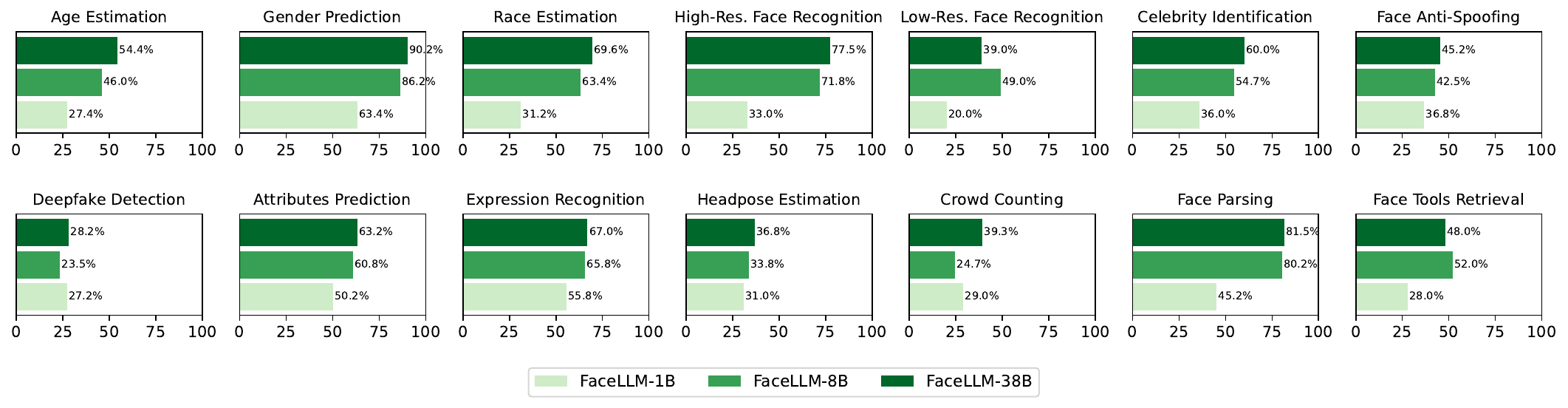}
    \caption{Performance of FaceLLM models (FaceLLM-1B, FaceLLM-8B, and FaceLLM-38B) on different sub-tasks, including age estimation, gender prediction, race estimation,
high-resolution face recognition, low-resolution face recognition, celebrity identification, face anti-spoofing, deepfake detection, 
attributes prediction, facial expression recognition, headpose estimation, 
face localization crowd counting, face parsing, 
and face tools retrieval.}
    \label{fig:facellm_performance}
\end{figure*}

As another experiment, we compare FaceLLM models with corresponding InternVL3~\cite{zhu2025internvl3} base models used for LoRA training with FairFaceGPT dataset. Table~\ref{tab:compare_base} compares the performance of FaceLLM models with base models for different tasks. As can be seen in this table, FaceLLM can improve the performance of base models (i.e., InternVL3) on some tasks while affecting the performance on some others. For example, we can see that for face recognition FaceLLM improves the performance for most versions of the InternVL3 models. However, for \textit{face tools use} task our finetuning leads to drop in the performance. In fact, the \textit{face tools use} task is not a face understanding task but rather it is a system design knowledge task, which includes \textit{text-only} questions  about face analysis tools\footnote{For example, this is a sample question-answer pair for \textit{face tools use} task: Question: \textit{in an airport security setting which should have face recognition and gender classification, which API call sequence should be used?} Answer: \textit{api\_7-identify\_high\_res\_face, api\_2-classify\_gender, api\_4-detect\_spoofing, api\_4-spoof\_confidence\_score, api\_5-detect\_deepfake}. } (i.e., no face image is given in the input). Since our training dataset (FairFaceGPT) does not have question-answer pairs for \textit{face tools use} task, we can expect that finetuning would reduce the performance in this task. 
Despite degradation of performance in individual tasks, we observe that all FaceLLM models improve the overall accuracy of their corresponding base models.

Figure~\ref{fig:facellm_performance} also compares the performance of different versions of FaceLLM  (1B, 8B, and 38B) on different sub-tasks in FaceXBench~\cite{narayan2025facexbench} including age estimation, gender prediction, race estimation,
high-resolution face recognition, low-resolution face recognition, celebrity identification, face anti-spoofing, deepfake detection, 
attributes prediction, facial expression recognition, headpose estimation, 
face localization crowd counting, face parsing, 
and face tools retrieval. 
As we can see, in most sub-tasks, FaceLLM-38B achieves the best performance. However, FaceLLM-8B has also a comparable performance with FaceLLM-38B, while having near 5 times less parameters.

\section{Conclusion}\label{section:Conclusion}
MLLMs are often trained on generic data and have limited capacity for domain-specific tasks, such as face analysis.
In this paper, we introduced \textbf{FaceLLM}, a multimodal large language model finetuned for facial image understanding, which achieves the state-of-the-art performance for different face-related tasks. To train FaceLLM, we proposed a weakly supervised pipeline based on attribute-aware prompting of ChatGPT to generate question-answer pairs. We used metadata in FairFace dataset to craft prompts which could guide ChatGPT to generate accurate answers, and then removed the attributes to form final questions for our dataset. The generated dataset, called \textbf{FairFaceGPT}, includes various face attributes including expression, pose, skin texture, and forensic information. 
FairFaceGPT can serve as a new resource for training MLLMs on face-centric vision-language tasks. Unlike traditional image captioning datasets, our dataset emphasizes structured and context-aware understanding of facial features.
Our results suggest that  supervision using LLMs offers a promising alternative to costly human annotation for vision-language tasks, particularly for sensitive data like facial analysis. We hope our research builds a foundation for future research in trustworthy, human-centric multimodal AI.

\section*{Acknowledgment}
This research is based upon work funded by the Hasler foundation through the Responsible Face Recognition (SAFER) project and the Swiss Center for Biometrics Research \& Testing at Idiap Research Institute.

{
\small
\bibliographystyle{ieeenat_fullname}
\bibliography{refs}
}

\end{document}